\documentclass{article} 
\usepackage[preprint]{colm2026_conference}

\usepackage{amsmath,amsfonts,bm}

\def\eqref#1{equation~\ref{#1}}

\def\1{\bm{1}}

\DeclareMathAlphabet{\mathsfit}{\encodingdefault}{\sfdefault}{m}{sl}
\SetMathAlphabet{\mathsfit}{bold}{\encodingdefault}{\sfdefault}{bx}{n}

\usepackage{hyperref}
\usepackage{url}
\usepackage{booktabs}
\usepackage{multirow}
\usepackage{graphicx}
\usepackage{subcaption}
\usepackage{amsmath,amssymb}

\title{Ex-GraphRAG: Interpretable Evidence Routing for Graph-Augmented LLMs}

\def\ours{{\textsc{Ex-GraphRAG}}}

\definecolor{cornflowerblue}{rgb}{0.39, 0.58, 0.93} 
\hypersetup{
    colorlinks=true,
    linkcolor=cornflowerblue,
    filecolor=magenta,      
    urlcolor=teal,
    citecolor=cornflowerblue,
    }

\author{\noindent
  Yoav Kor Sade\thanks{Equal contribution.} \\
  Tel Aviv University \\
  \And
  Arvindh Arun\footnotemark[1] \\
  Institute for AI, University of Stuttgart \\
  \And
  Rishi Puri \\
  NVIDIA \\
  \And
  Steffen Staab \\
  Institute for AI, University of Stuttgart \\
  \And
  Maya Bechler-Speicher \\
  Meta AI\\
}

\begin{document}
\maketitle

\begin{abstract}
GraphRAG conditions language models on subgraphs retrieved from knowledge graphs, encoded via message-passing GNNs. Because these encoders entangle node contributions through iterated neighborhood aggregation, there is no closed-form way to determine how much each retrieved entity influenced the encoder's output, and therefore no way to faithfully audit what structural evidence actually reached the model. We introduce Ex-GraphRAG, which replaces the GNN encoder with a Multivariate Graph Neural Additive Network (M-GNAN), an extension of additive graph models to high-dimensional embedding spaces that yields an exact decomposition of the encoder's output across individual nodes and feature groups, without post-hoc approximation. On STaRK-Prime, this auditable encoder matches black-box performance. Using it to audit evidence routing, we uncover a semantic-structural mismatch: the nodes that dominate the encoder's output are structurally disconnected in the retrieved subgraph, held together by low-attribution intermediaries whose removal degrades multi-hop QA by up to 28\%. This mismatch, invisible to any opaque encoder, reveals that semantic importance and structural connectivity are governed by disjoint sets of nodes, with direct implications for retrieval pruning, context construction, and failure diagnosis in graph-augmented LLMs.

\end{abstract}

\section{Introduction}

Retrieval-Augmented Generation (RAG) improves the factuality of Large Language Models (LLMs) by conditioning generation on external knowledge retrieved at inference time~\citep{lewis2020retrieval, gao2024rag}. GraphRAG extends this to structured knowledge: given a natural-language query, a query-specific subgraph is retrieved from a knowledge graph, encoded with a Graph Neural Network (GNN), and projected into the LLM's input space as a soft prompt~\citep{gret, han2025retrievalaugmentedgenerationgraphsgraphrag}. The GNN encoder acts as an information bottleneck, compressing hundreds of retrieved entities and relations into latent vectors. Standard GNN encoders like GCN~\citep{kipf2017semi}, GAT~\citep{velickovic2018graph}, and GIN~\citep{xu2019powerful} have repeated neighborhood aggregation followed by nonlinear transformations across multiple layers, producing representations that entangle the contributions of each node with those of its multi-hop neighborhood. The result is that prompt vectors fed to the LLM have no closed-form per-node decomposition, making it impossible to faithfully audit which retrieved entities actually influenced the model's conditioning.

Auditing a GraphRAG pipeline requires knowing what the encoder did: which nodes it weighted, which it suppressed, and whether the graph-conditioned prompt reflects the relevant structural evidence. When the system produces an incorrect answer, the failure can originate at the retriever, the encoder, or the LLM, and disentangling these requires per-node attribution at the encoder level. In high-stakes settings such as biomedical QA and regulatory compliance, this kind of auditing is not optional~\citep{rudin2019stop}. Post-hoc explainability methods can be applied to the encoder: GNNExplainer~\citep{gnnexplainer} optimizes a mutual information objective, SubgraphX~\citep{yuan2021subgraphx} uses Shapley-value-based search, and attention weights from GAT layers are sometimes treated as importance scores. But these provide approximations whose faithfulness to the model's actual computation is not guaranteed~\citep{yuan2023explainability}. Attention weights in particular correlate poorly with the features that causally influence predictions~\citep{jain2019attention, wiegreffe2019attention}. For faithful auditing, approximate methods are insufficient.

To make the graph encoder auditable, we replace it with a Multivariate Graph Neural Additive Network (M-GNAN). Building on GNAN~\citep{bechler-speicher2024the}, which decomposes graph-level predictions additively across nodes and features by architectural design, M-GNAN extends the formalism to \emph{feature groups}, accommodating the high-dimensional embedding vectors used in GraphRAG systems where individual dimensions carry no independent semantic meaning. We call the resulting framework \ours, whose importance scores are exact partial sums of the encoder's output rather than post-hoc approximations.

We audit evidence routing on a biomedical KGQA dataset using \ours, to explicitly quantify the \emph{semantic-structural mismatch} in the retrieved subgraphs: we demonstrate precisely how the nodes that dominate the encoder's output are typically poorly connected to one another. Low-importance intermediary nodes (drug classes, pathway identifiers, shared protein families) serve as structural bridges between them, and pruning these bridges degrades multi-hop QA by up to 28\%. This mismatch between semantic importance and structural connectivity is invisible to opaque encoders and would be obscured by post-hoc approximations. Our main contributions are:
\begin{itemize}
    \item We audit evidence routing in GraphRAG and uncover a \textit{semantic-structural mismatch}: the nodes that dominate the encoder's output are structurally disconnected, connected through low-attribution intermediaries whose removal degrades multi-hop reasoning by up to 28\%.
    \item To enable this audit, we introduce \textit{M-GNAN}, extending GNAN to feature-group embeddings, and integrate it into the G-Retriever framework as an intrinsically decomposable encoder (\ours).
    \item On STaRK-Prime, \ours\ matches the QA performance of black-box encoders (GAT, GCN, GIN) while producing exact node-level attribution.
    \item We characterize additional capabilities that auditable encoders enable: retrieval debugging, importance-guided context construction, and multi-granularity attribution.
\end{itemize}

\section{Related Work}

\subsection{Graph Retrieval-Augmented Generation}

GraphRAG systems integrate structured knowledge from graphs into LLM generation pipelines. G-Retriever~\citep{gret} retrieves query-specific subgraphs via Prize-Collecting Steiner Trees, encoding them with a GNN, and projecting the resulting representations into the LLM's input space. Subsequent work has explored alternative graph-text integration strategies~\citep{perozzi2024letgraphtalkingencoding}, community-level graph summarization for broader query coverage~\citep{edge2024local}, GNN-based dense retrieval for LLM reasoning~\citep{mavromatis2025gnnrag}, relation-path-grounded reasoning for interpretable KG-QA~\citep{luo2024rog}, and some works covering the broader integration of LLMs with knowledge graphs \citep{pan2024unifying}. Earlier work on GNN-augmented QA over knowledge graphs, notably QA-GNN~\citep{yasunaga2021qagnn}, demonstrated the value of jointly reasoning over language context and graph structure, but did not address the transparency of the graph encoder itself. Surveys of the GraphRAG landscape~\citep{han2025retrievalaugmentedgenerationgraphsgraphrag} identify the GNN encoder as a central but underexamined component: while retrieval strategies and LLM architectures receive significant attention, the encoder is typically treated as an interchangeable module evaluated only by downstream task performance. Several of these systems offer partial transparency at other stages: G-Retriever returns the retrieved subgraph, giving visibility into \emph{what} was retrieved but not how the encoder weighted it; GNN-RAG~\citep{mavromatis2025gnnrag} assigns importance scores to nodes for retrieval ranking, but these scores guide subgraph selection, not encoder auditing; RoG~\citep{luo2024rog} grounds reasoning in relation paths, sidestepping the encoder entirely. None make the encoder itself transparent. Our work addresses this gap: we replace the encoder with an intrinsically decomposable architecture so that per-node attribution is exact and available for auditing.

\subsection{Explainability for Graph Neural Networks}

Explainability methods for GNNs fall broadly into post-hoc and intrinsic categories~\citep{yuan2023explainability}. Post-hoc methods explain a trained model after the fact: GNNExplainer~\citep{gnnexplainer} identifies important substructures by maximizing mutual information between explanation and prediction; PGExplainer~\citep{luo2020parameterized} amortizes this via a learned mask generator; SubgraphX~\citep{yuan2021subgraphx} searches for explanatory subgraphs using Monte Carlo tree search with Shapley values; gradient-based and perturbation-based approaches compute attribution scores through backpropagation or input masking. Attention-based attribution interprets the learned coefficients of GAT-style models as importance, though this interpretation is contested. \citet{jain2019attention} showed that attention weights frequently fail to correlate with gradient-based importance, and \citet{brody2022how} demonstrated that standard GAT computes a limited form of static attention that cannot fully rank neighbor importance. \citet{wiegreffe2019attention} argued that attention can carry explanatory signal under certain conditions, but the consensus remains that attention weights are unreliable as faithful attributions~\citep{yuan2023explainability}.

Intrinsic methods instead constrain the architecture so that explanations follow directly from the computation. This tradition extends from Generalized Additive Models~\citep{hastie1990generalized}, which decompose predictions as sums of univariate functions, through gradient-boosted GAMs~\citep{lou2012intelligible, lou2013accurate} and Neural Additive Models~\citep{agarwal2021neural}, to GNAN~\citep{bechler-speicher2024the}, which brings the additive principle to graph-structured data. Unlike post-hoc methods such as SHAP~\citep{lundberg2017shap}, which estimate per-feature contributions by sampling perturbations of a black-box model, intrinsic additive models produce importance scores that are the literal terms in the summation constituting the output, without any estimation or approximation involved. To our knowledge, no prior work has applied an intrinsically decomposable GNN as the encoder in a GraphRAG system.

\section{Methodology}


The standard GraphRAG pipeline uses a message-passing GNN (GCN, GAT, or GIN) to encode the retrieved subgraph into node representations, which are then aggregated and projected into the LLM's embedding space. These encoders interleave neighborhood aggregation with nonlinear transformations across $L$ layers, producing representations where the contribution of any single input node is entangled with the contributions of its multi-hop neighborhood. The output has no closed-form per-node decomposition.

GNAN~\citep{bechler-speicher2024the} takes a different architectural approach. Rather than entangling node contributions through message passing, GNAN computes each node's contribution to the output independently through learned \emph{shape functions}, then combines contributions additively with distance-based weighting. The result is a graph-level prediction that decomposes exactly into a sum over nodes (and over features within each node), where each term can be computed and inspected independently. This additive structure is what makes the encoder auditable: given its output, we can read off how much each retrieved node contributed, without approximation.

The \ours\ pipeline proceeds as follows: (1) embedding the query, retrieving seed nodes by vector similarity and expanding seeds into a subgraph via PCST (Section~\ref{sec:subgraph}), (2) encoding the subgraph with M-GNAN, producing node representations alongside per-node importance scores (Section~\ref{sec:mgnan}), and (3) projecting the representations into the LLM's input space (Section~\ref{sec:projection}). The importance scores are not computed separately, but they are partial sums of the encoder's output, extracted directly from the forward pass.

\subsection{Graph Construction for Knowledge Graph Question Answering}
\label{sec:subgraph}

Given a natural-language question, we construct a question-specific subgraph that serves as the encoder's input. Both pipelines below start from the same seed selection: we embed the question with a sentence transformer and retrieve the top-$k=4$ seed nodes by cosine similarity from the knowledge graph's node embeddings. They differ in how they expand from these seeds.

\paragraph{Single-hop PCST-merge.}
From the seed nodes, we perform a 1-hop expansion to form a candidate pool. We then run the Prize-Collecting Steiner Tree algorithm, assigning prizes to the top-100 nodes by query similarity and using uniform edge costs. To ensure high recall of semantically relevant entities, we merge the PCST output with the induced subgraph of the top-200 most similar nodes from the candidate pool.

\paragraph{Multi-hop PCST-merge.}
To capture longer-range dependencies, we expand from seeds for $h=3$ hops, pruning the frontier at each step by retaining only the top-$k_{\text{seed}}=5$ neighbors per node (ranked by query similarity) to manage graph size. After expansion, we apply the same PCST-plus-merge procedure. The multi-hop pipeline produces larger, more structurally complex subgraphs that can capture reasoning chains spanning several entity types, at the cost of including more noise and more intermediary nodes. As we show in Section~\ref{sec:analysis}, these intermediaries turn out to play a critical structural role.

\subsection{Extension of GNAN to Node Embedding Vectors (M-GNAN)}
\label{sec:mgnan}

GNAN is defined over per-feature shape functions: each scalar feature has its own univariate function, and importance is attributed per-feature. But GraphRAG nodes are typically represented by high-dimensional embedding vectors (e.g., 768 or 1536 dimensions from sentence transformers) whose individual coordinates lack independent semantic meaning. Attributing importance to, say, dimension 437 of a sentence embedding is uninterpretable. To bridge this gap, we extend GNAN to operate on \emph{feature groups}: subsets of the feature vector that are treated as a unit for both computation and attribution. We call this extension \emph{Multivariate GNAN} (M-GNAN).

We partition the feature space into $G$ groups, where each group $g \in \{1, \ldots, G\}$ contains a subset of feature indices $\mathcal{F}_g \subseteq \{1, \ldots, d\}$. In the common GraphRAG setting of a single embedding vector per node, we set $G=1$ and $\mathcal{F}_1=\{1,\ldots,d\}$, so that the entire embedding is one group. For each feature group $g$, we define a \emph{multivariate shape function}:
\[
  f_g(\mathbf{x}_g; \theta_g) : \mathbb{R}^{|\mathcal{F}_g|} \to \mathbb{R},
\qquad
  \mathbf{x}_g = [x_{k_1}, \ldots, x_{k_{|\mathcal{F}_g|}}], \ k_i \in \mathcal{F}_g,
\]
parameterized as a multi-layer perceptron. Note that while the shape function is a flexible nonlinear function of its input group, the overall model remains additive across groups and nodes. The expressiveness constraint is on how nodes \emph{combine}, not on how individual node features are processed. The representation for node $i$ in feature group $g$ is:
\[
  \left[\mathbf{h}_i\right]_g
  = \sum_{j=1}^{N}
    \frac{1}{\#\text{dist}(j, i)} \cdot
    \rho\!\left(\frac{1}{1 + \text{dist}(j, i)}\right) \cdot
    f_g\!\left([\mathbf{x}_j]_{\mathcal{F}_g}\right),
\]
where $\text{dist}(j,i)$ is the shortest-path distance between nodes $j$ and $i$ in the subgraph, $\#\text{dist}(j,i)$ is the number of nodes at the same distance from $i$ as $j$ (a normalization factor that prevents distant, well-connected neighborhoods from dominating), and $\rho$ is a learnable monotone function that controls how influence decays with graph distance. For graph-level prediction, the output decomposes as:
\begin{align*}
  \sigma\left(\sum_{g=1}^{G} \sum_{i=1}^{N} [\mathbf{h}_i]_g\right)
  &= \sigma\left(\sum_{g=1}^{G} \sum_{j=1}^{N} f_g\!\left([\mathbf{x}_j]_{\mathcal{F}_g}\right)
    \sum_{i=1}^{N} \frac{1}{\#\text{dist}(j, i)} \cdot
    \rho\!\left(\frac{1}{1 + \text{dist}(j, i)}\right)\right),
\end{align*}
which provides an exact additive decomposition over both nodes and feature groups. The importance of node $j$ in group $g$ is the corresponding inner term. It is not estimated, but computed directly from the model's forward pass.

\subsection{Projection and Generation}
\label{sec:projection}

Following~\citet{gret}, the LLM receives two versions of the retrieved subgraph. First, M-GNAN's node representations are mapped to the LLM's hidden dimension via a learned MLP, producing a graph token that is prepended to the input as a soft prompt. Second, the subgraph is linearized into text by flattening node and edge descriptions, and concatenated with the query as standard textual input. The LLM attends to both the graph-conditioned soft prompt and the textualized subgraph during generation. We fine-tune the LLM with LoRA~\citep{hu2022lora} while jointly training the M-GNAN encoder and projection layer end-to-end.

\section{What Auditable Graph Encoding Enables}
\label{sec:capabilities}

An auditable encoder opens capabilities that opaque encoders cannot support. Because M-GNAN's output decomposes additively across nodes, per-node importance scores are available at inference time as a byproduct of encoding, without additional forward passes or post-hoc analysis. We demonstrate two of these capabilities in Sections~\ref{sec:experiments} and~\ref{sec:analysis}; we briefly note additional directions the formalism supports at the end of this section.

\subsection{Auditing and Debugging Retrieval Pipelines}

When a GraphRAG system produces an incorrect answer, the failure can originate at any pipeline stage: the query embedding may miss relevant seed nodes, the PCST expansion may select a poorly connected subgraph, the GNN encoder may misweight the evidence, or the LLM may ignore the graph-conditioned prompt entirely. With opaque encoders, disentangling retrieval failures from encoding failures is difficult. \ours's importance decomposition enables targeted analysis: if importance concentrates on query-relevant entities but the LLM still makes an error, the encoder is functioning correctly and the problem lies downstream. If importance concentrates on irrelevant nodes, the encoder's training or the subgraph's composition is at fault. Stage-wise debugging follows directly from exact per-node attribution.

\subsection{Disentangling Semantic and Structural Importance}

\ours's additive decomposition quantifies each node's direct semantic contribution to the encoder's output. But as we show in Section~\ref{sec:analysis}, semantic importance is not the only dimension along which nodes matter: structural importance, the role a node plays in connecting other nodes, is distinct and sometimes conflicting. We term this divergence the \emph{semantic-structural mismatch}. Combining \ours's semantic scores with graph-theoretic structural measures (e.g., betweenness centrality) could produce composite importance maps that classify nodes along both dimensions. We examine the mismatch empirically in Section~\ref{sec:analysis}.

\paragraph{Further capabilities.} The exact decomposition supports additional applications that we do not evaluate in this work. On the context construction side, node importance scores could guide the linearization order of subgraph descriptions passed to the LLM, placing high-importance nodes earlier to mitigate known positional biases in long-context attention~\citep{liu2024lost}; our PCST\,+\,Top-25 configuration (Section~\ref{sec:experiments}) is a simple instance of this idea. On the attribution side, M-GNAN's formalism generalizes to multiple feature groups ($G > 1$), allowing separate attribution of importance, e.g., to a node's textual embedding versus its molecular fingerprint. We leave both of these as potential directions for future work to explore.

\section{Empirical Evaluation}
\label{sec:experiments}

The capabilities above rest on an assumption that M-GNAN can replace standard GNN encoders without substantial loss in downstream QA performance. We evaluate this on knowledge graph question answering, focusing on two questions: (1) does M-GNAN maintain competitive performance relative to black-box encoders? and (2) what happens when its importance scores are used to filter or emphasize the context passed to the LLM?

\paragraph{Dataset.}
STaRK-Prime~\citep{wu2024stark}, a knowledge graph QA benchmark built on a biomedical knowledge graph. We use the standard train/val/test splits.

\paragraph{Baselines.}
G-Retriever with GAT, GCN, and GIN encoders using PCST-based subgraph retrieval, following the same setup as \citet{gret}. Llama 3.1 8B~\citep{grattafiori2024llama3herdmodels} is used as the LLM backbone for all configurations.

\paragraph{Context configurations.}
For each retrieval pipeline, we evaluate three configurations using M-GNAN:
\begin{itemize}
    \item \textbf{PCST:} Full PCST subgraph linearized and passed to the LLM (standard).
    \item \textbf{Top-25 only:} Only the top-25 nodes by M-GNAN importance score, discarding the rest.
    \item \textbf{PCST + Top-25:} Full PCST subgraph with the top-25 important nodes emphasized via ordering and special markers in the text context.
\end{itemize}

\paragraph{Metrics.}
Hits@1, Hits@5, F1 score, and Mean Reciprocal Rank (MRR).

\begin{table}[t]
\centering
\caption{\textbf{Single-hop PCST-merge retrieval.} M-GNAN provides exact node-level attribution while maintaining competitive performance. Best results in \textbf{bold}.}
\label{tab:singlehop}
\begin{tabular}{llcccc}
\toprule
\textbf{GNN} & \textbf{Context} & \textbf{Hits@1} & \textbf{Hits@5} & \textbf{F1} & \textbf{MRR} \\
\midrule
\multicolumn{6}{l}{\textit{Black-box baselines}} \\
GAT & PCST & \textbf{0.297} & \textbf{0.335} & \textbf{0.243} & \textbf{0.313} \\
GCN & PCST & 0.291 & 0.331 & 0.239 & 0.308 \\
GIN & PCST & 0.292 & 0.336 & 0.239 & 0.310 \\
\midrule
\multicolumn{6}{l}{\textit{Auditable (Ours)}} \\
M-GNAN & PCST & 0.285 & 0.335 & 0.236 & 0.306 \\
M-GNAN & PCST + Top-25 & 0.289 & 0.332 & 0.237 & 0.307 \\
M-GNAN & Top-25 only & 0.257 & 0.306 & 0.220 & 0.279 \\
\bottomrule
\end{tabular}
\end{table}

\begin{table}[t]
\centering
\caption{\textbf{Multi-hop PCST-merge retrieval.} In the more challenging multi-hop setting, M-GNAN remains competitive with GAT and GCN but trails GIN, reflecting the expressiveness--auditability tradeoff. Best results in \textbf{bold}.}
\label{tab:multihop}
\begin{tabular}{llcccc}
\toprule
\textbf{GNN} & \textbf{Context} & \textbf{Hits@1} & \textbf{Hits@5} & \textbf{F1} & \textbf{MRR} \\
\midrule
\multicolumn{6}{l}{\textit{Black-box baselines}} \\
GAT & PCST & 0.282 & 0.331 & 0.231 & 0.302 \\
GCN & PCST & 0.291 & 0.331 & 0.235 & 0.309 \\
GIN & PCST & \textbf{0.341} & \textbf{0.387} & \textbf{0.291} & \textbf{0.361} \\
\midrule
\multicolumn{6}{l}{\textit{Auditable (Ours)}} \\
M-GNAN & PCST & 0.293 & 0.337 & 0.239 & 0.312 \\
M-GNAN & PCST + Top-25 & 0.283 & 0.330 & 0.234 & 0.303 \\
M-GNAN & Top-25 only & 0.212 & 0.253 & 0.175 & 0.229 \\
\bottomrule
\end{tabular}
\end{table}

\paragraph{Results.}
In the single-hop setting (Table~\ref{tab:singlehop}), M-GNAN with the full PCST context achieves 0.285 Hits@1, within 0.012 of the best baseline (GAT, 0.297), and matches the baselines on Hits@5 (0.335). The gap is small and consistent across metrics, M-GNAN pays a modest cost for exact decomposability.

The multi-hop setting (Table~\ref{tab:multihop}) is more mixed. M-GNAN (0.293 Hits@1) surpasses GAT (0.282) and matches GCN (0.291), but trails GIN (0.341). This gap is expected. GIN is maximally expressive among message-passing GNNs in the WL hierarchy~\citep{xu2019powerful}, and that expressiveness matters more in larger multi-hop subgraphs with more complex structure. M-GNAN's additive constraint, the property that enables exact decomposition, limits its capacity to model certain interaction effects between nodes. This is the explicit tradeoff: M-GNAN gives up expressiveness in the hardest setting for guarantees about its importance scores.

\paragraph{Pruning less important nodes.}
Filtering context to only the top-25 nodes by importance (Top-25 only) degrades performance substantially, as shown in Table~\ref{tab:ablation}. Hits@1 drops from 0.285 to 0.257 ($-9.8\%$) in single-hop and from 0.293 to 0.212 ($-27.6\%$) in multi-hop. The multi-hop degradation is nearly three times worse. This asymmetry reflects the structure of multi-hop subgraphs: they are larger, contain more intermediary entities, and depend more on structural bridges between answer-relevant nodes. The emphasis configuration (PCST + Top-25) recovers performance in the single-hop setting (0.289) but only partially in multi-hop (0.283). The next section examines why through a closer look at importance distributions.

\begin{table}[t]
\centering
\caption{\textbf{Effect of context configuration on M-GNAN performance.} Aggressive pruning (Top-25 only) degrades performance substantially, with the effect nearly three times worse in multi-hop settings. The emphasis configuration (PCST + Top-25) partially recovers performance by retaining structural context while highlighting important nodes.}
\label{tab:ablation}
\begin{tabular}{llcccc}
\toprule
\textbf{Retrieval} & \textbf{Context} & \textbf{Hits@1} & \textbf{$\Delta$Hits@1} & \textbf{F1} & \textbf{$\Delta$F1} \\
\midrule
\multirow{3}{*}{Single-hop}
& PCST & 0.285 & --- & 0.236 & --- \\
& PCST + Top-25 & 0.289 & +1.4\% & 0.237 & +0.4\% \\
& Top-25 only & 0.257 & $-$9.8\% & 0.220 & $-$6.8\% \\
\midrule
\multirow{3}{*}{Multi-hop}
& PCST & 0.293 & --- & 0.239 & --- \\
& PCST + Top-25 & 0.283 & $-$3.4\% & 0.234 & $-$2.1\% \\
& Top-25 only & 0.212 & $-$27.6\% & 0.175 & $-$26.8\% \\
\bottomrule
\end{tabular}
\end{table}

\section{Auditing Evidence Routing}
\label{sec:analysis}

The pruning results raise a natural question: \textit{if the top-scoring nodes account for most of the encoder's output, why does removing the rest hurt performance so badly?} We use M-GNAN's exact decomposition to examine this on a representative query.

\subsection{Node Importance Distributions}

\begin{figure}[t]
    \centering
    \includegraphics[width=0.95\linewidth]{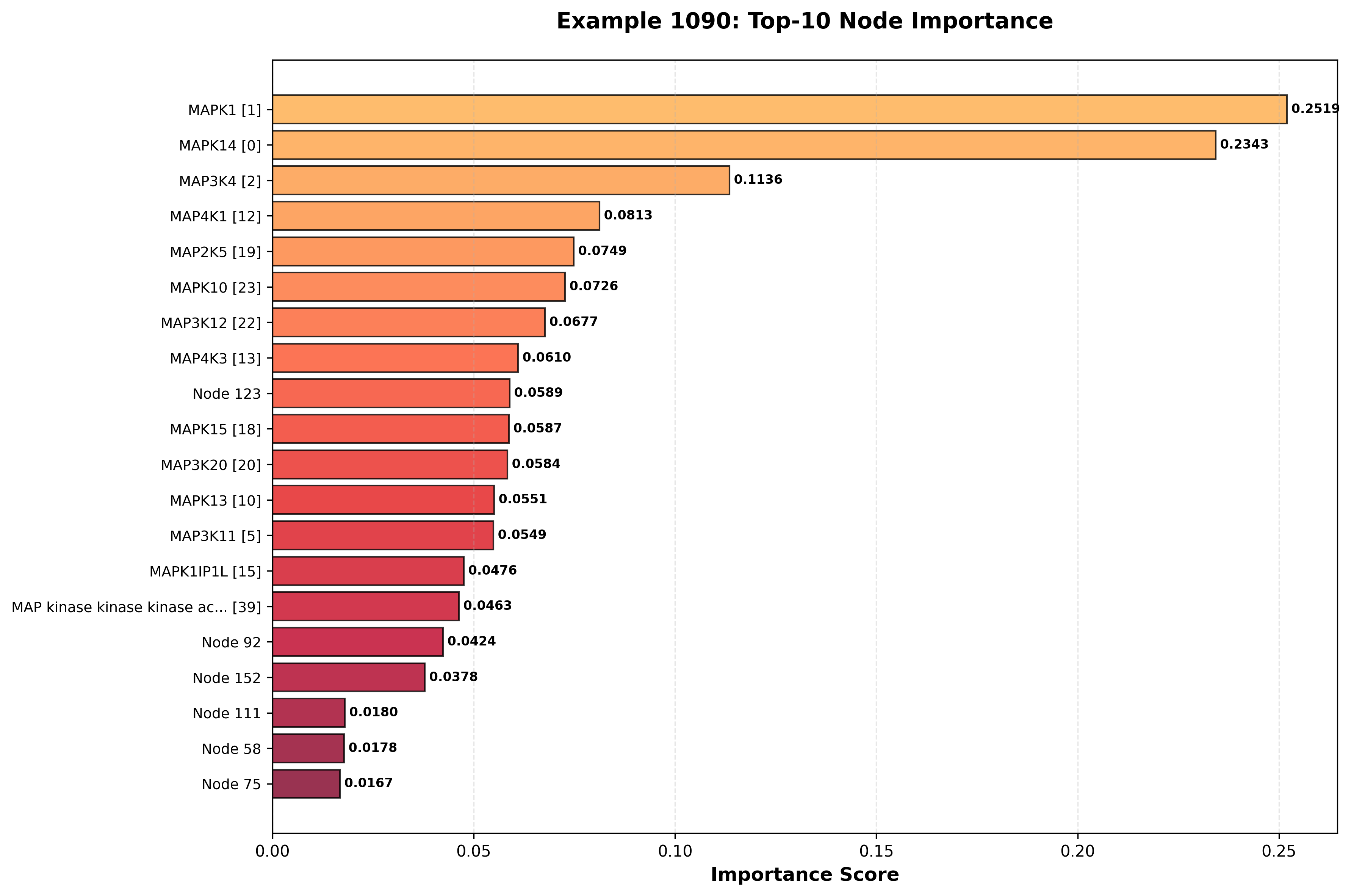}
    \caption{\textbf{Node importance scores for a representative query.} M-GNAN importance scores for the query \textit{``Could you list the tablet or capsule medications that act on the MAPK1 gene/protein?''} Three nodes (MAPK1, MAPK14, MAP3K4) account for approximately 60\% of total importance; the remaining 180 nodes each contribute less than 0.05. Yet removing these low-scoring nodes causes 28\% performance degradation in multi-hop settings (Table~\ref{tab:multihop}).}
    \label{fig:importance_bars}
\end{figure}

Figure~\ref{fig:importance_bars} shows M-GNAN's importance scores for a representative query from STaRK-Prime: \textit{``Could you list the tablet or capsule medications that act on the MAPK1 gene/protein?''} Importance distribution is very skewed, with just three nodes: MAPK1 (0.252), MAPK14 (0.234), and MAP3K4 (0.114), which account for roughly 60\% of total importance. After rank 10, scores drop below 0.05; the vast majority of the 200 nodes contribute less than 0.02 each.

The concentration aligns with domain knowledge: MAPK1 is the query target, MAPK14 is a closely related kinase, and MAP3K4 is an upstream activator in the same signaling cascade. In a subgraph spanning hundreds of nodes across multiple entity types (drugs, proteins, pathways, diseases), the encoder's output is driven by a small semantic core. However, semantic irrelevance does not imply structural irrelevance. Removing these low-scoring nodes degrades QA performance by up to 28\% (Section~\ref{sec:experiments}), with more details in Figure~\ref{fig:subgraph}.

\subsection{Structural Fragmentation Under Pruning}

\begin{figure}[t]
    \centering
    \includegraphics[width=0.95\linewidth]{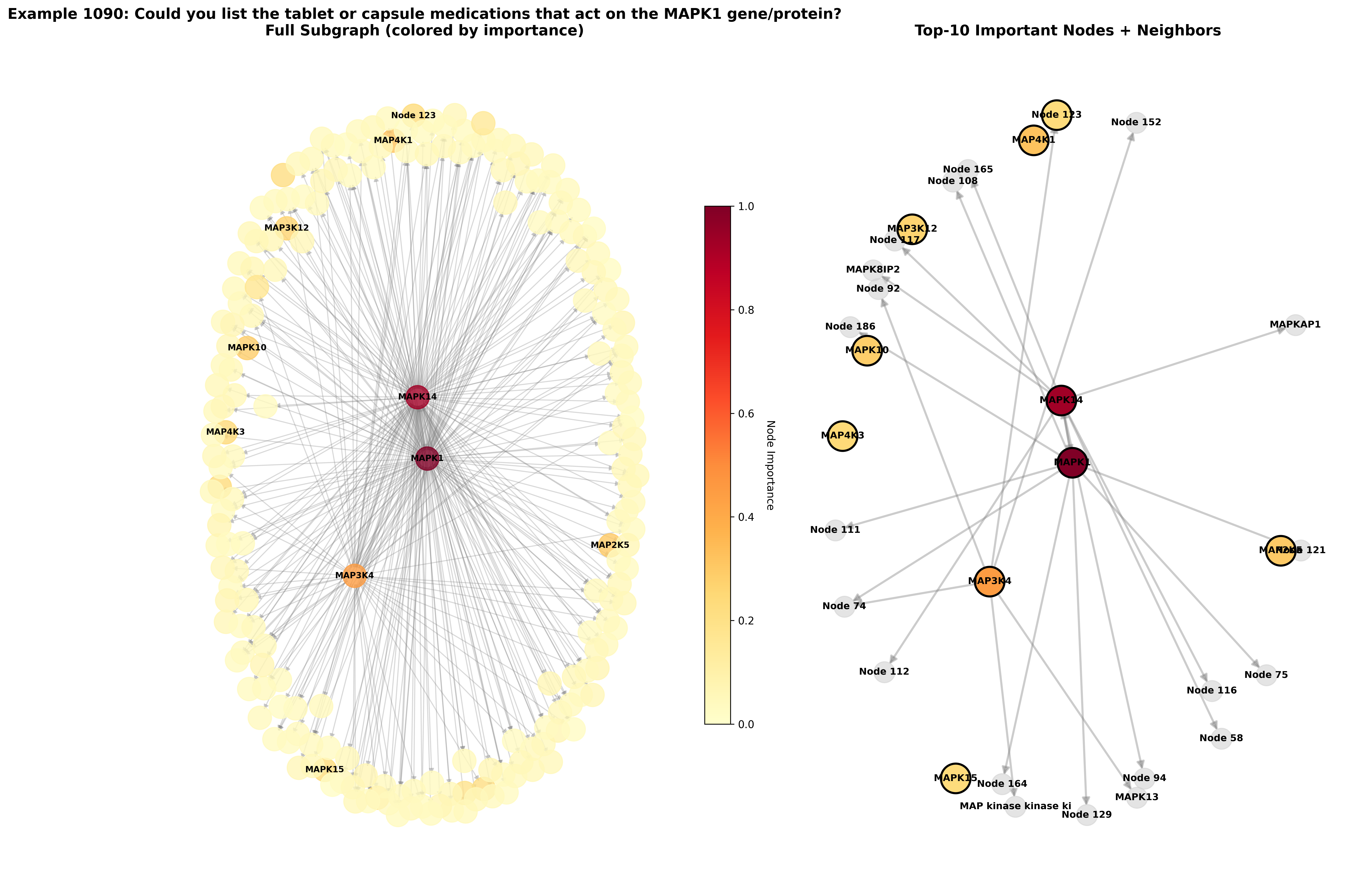}
    \caption{\textbf{Full subgraph vs.\ induced subgraph on high-importance nodes.} Left: the full retrieved subgraph for the MAPK1 query, colored by M-GNAN importance (dark red = high, yellow = low). High-importance nodes sit within a dense network of low-importance intermediaries. Right: the induced subgraph on the top-10 important nodes and their immediate neighbors. Removing bridge nodes fragments the graph into disconnected components.}
    \label{fig:subgraph}
\end{figure}

Figure~\ref{fig:subgraph} visualizes the same query's subgraph in two views. The left panel shows the full retrieved subgraph colored by importance: high-importance nodes (dark red) sit within a larger network of low-importance intermediaries (yellow). The right panel restricts to the top-10 important nodes and their immediate neighbors. The graph fragments into disconnected components. Nodes that were close together in the full subgraph become distant once intermediaries are removed.

We term these \emph{bridge nodes}: entities with negligible M-GNAN importance that maintain connectivity between high-importance nodes. In the MAPK1 example, bridges include intermediate drug classes connecting medications to target proteins, pathway identifiers linking kinases within the MAPK cascade, and shared protein families providing categorical links between otherwise unrelated entities. None are answers to the query, and M-GNAN correctly assigns them low importance. But removing them fragments the graph, destroying the relational context the LLM needs for reasoning.

The 28\% Hits@1 degradation in multi-hop pruning (Table~\ref{tab:multihop}) follows from this: multi-hop subgraphs span more entity types and longer reasoning chains, requiring more bridges. Without them, the LLM receives disconnected entity descriptions rather than a connected subgraph.

\subsection{The Semantic-Structural Mismatch}
Semantic relevance and structural connectivity divergence in knowledge graphs is not a new observation \cite{yuan2023explainability,edge2024local}; what has been lacking is the ability to quantify this divergence faithfully within a specific encoder, on a per-query basis. M-GNAN provides exactly this: turning a qualitative intuition into a per-node, per-query diagnostic without post-hoc approximation. M-GNAN's scores measure each node's direct additive contribution to the encoder's output, i.e., its semantic importance. But downstream reasoning also depends on structural connectivity, a global property that no per-node score can capture. A node can be indispensable for reasoning (its removal disconnects the graph) while contributing almost nothing to the output in isolation.

Any node-level attribution method faces this mismatch~\citep{yuan2023explainability}. Exact decomposition does not resolve it, but it does surface it. With a post-hoc method, bridge nodes might receive moderate importance scores, obscuring whether they genuinely contribute to the computation or just serve as neighbors to nodes that do. M-GNAN's exact scores make the bimodal structure of the importance landscape clear: a small core of high-attribution semantic nodes, surrounded by low-attribution structural bridges.

Combining semantic importance from M-GNAN with graph-theoretic structural measures (e.g., betweenness centrality) could yield composite scores that account for both dimensions, enabling principled pruning that retains both the semantic core and the structural bridges. This connects to the multi-granularity capability described in Section~\ref{sec:capabilities}.

\section{Discussion}

\paragraph{The expressiveness-auditability tradeoff.}
Our results make the tradeoff between expressiveness and interpretability concrete. In the single-hop setting, M-GNAN's performance cost relative to black-box encoders is small (0.285 vs.\ 0.297 Hits@1 for GAT). In multi-hop, the gap widens against GIN (0.293 vs.\ 0.341), though M-GNAN still matches or exceeds GAT and GCN. The GIN gap reflects the architectural cost of additive decomposability (see Section~\ref{sec:experiments}). When faithful attribution is a requirement~\citep{rudin2019stop}, this cost is justified; when maximum accuracy is the sole priority, GIN with post-hoc attribution remains a reasonable alternative. The semantic-structural mismatch (Section~\ref{sec:analysis}) is one example of the insights this transparency makes accessible; while the theoretical divergence between structural and semantic node roles is a known concept in graph learning, \ours\ provides the mechanism to reliably audit and isolate this behavior in production retrieval pipelines. Retrieval debugging, importance-guided context construction, and multi-granularity attribution are other areas that remain to be explored empirically.

\paragraph{Limitations.}
We evaluate on a single dataset (STaRK-Prime), and our quantitative findings may not generalize to other knowledge graphs, question types, or LLM backbones. The audit covers one query in detail; broader analysis of the mismatch across queries and graph domains would strengthen the finding. We do not compare M-GNAN's importance scores against post-hoc attributions applied to the same pipeline, report error bars across training runs, or conduct user studies on whether importance scores improve human verification of LLM answers.

\paragraph{Future work.}
The most immediate extensions are multi-dataset evaluation (STaRK-MAG and other KGQA benchmarks), empirical comparisons with post-hoc explainability methods using the same G-Retriever pipeline, and composite importance scores that combine M-GNAN's semantic attribution with graph-theoretic structural measures. User studies measuring whether practitioners can diagnose retrieval failures faster with \ours\ than with opaque baselines would validate the practical utility of the approach.

\section{Conclusion}

We audited evidence routing in GraphRAG and uncovered a semantic-structural mismatch: the nodes most important to the encoder's output are structurally disconnected, held together by low-attribution intermediaries whose removal degrades multi-hop reasoning by up to 28\%. This finding was made possible by \ours, which replaces the GNN encoder with M-GNAN, an intrinsically decomposable architecture that yields exact per-node attribution at competitive performance. The mismatch is invisible to opaque encoders and would be obscured by post-hoc approximations that blur the boundary between semantic and structural roles. As GraphRAG systems are deployed in high-stakes settings, the ability to audit the encoder through tools like \ours, not just the retriever or the LLM, will be increasingly important.

\bibliography{main}
\bibliographystyle{plainnat}

\end{document}